\documentclass{article}
\usepackage{ijcai25}

\usepackage{times}
\usepackage{soul}
\usepackage{url}
\usepackage[hidelinks]{hyperref}
\usepackage[utf8]{inputenc}
\usepackage[small]{caption}
\usepackage{graphicx}
\usepackage{amssymb}
\usepackage{amsmath}
\usepackage{amsthm}
\usepackage{booktabs}
\usepackage{algorithm}
\usepackage{algorithmic}
\usepackage{xspace}
\usepackage[switch]{lineno}

\usepackage{multirow}
\usepackage{xcolor}

\title{Dynamic and Adaptive Feature Generation with LLM}

\author{
Xinhao Zhang$^1$
\and
Jinghan Zhang$^1$
\and
Banafsheh Rekabdar$^1$
\and
Yuanchun Zhou$^2$
\and
Pengfei Wang$^{2,3}$\thanks{Corresponding author.} 
\And 
Kunpeng Liu$^1$\\ 
\affiliations
$^1$Portland State University\\
$^2$Computer Network Information Center, Chinese Academy of Sciences\\
$^3$University of Chinese Academy of Sciences, Chinese Academy of Sciences\\ 
\emails
\{xinhaoz, jinghanz, kunpeng, rekabdar\}@pdx.edu,
\{zyc, pfwang\}@cnic.cn
}

\newcommand{\model}{LFG\xspace}

\definecolor{bgcolor1}{HTML}{DAE8FC}
\definecolor{bgcolor2}{HTML}{D5E8D4}
\definecolor{bgcolor3}{HTML}{E1D5E7}

\definecolor{txtcolor1}{HTML}{6C8EBF}
\DeclareMathOperator*{\argmax}{arg\,max}

\begin{document}

\maketitle

\begin{abstract}
    The representation of feature space is a crucial environment where data points get vectorized and embedded for subsequent modeling. Thus, the efficacy of machine learning (ML) algorithms is closely related to the quality of feature engineering. As one of the most important techniques, feature generation transforms raw data into an optimized feature space conducive to model training and further refines the space. Despite the advancements in automated feature engineering and feature generation, current methodologies often suffer from three fundamental issues: lack of explainability, limited applicability, and inflexible strategy. These shortcomings frequently hinder and limit the deployment of ML models across varied scenarios. Our research introduces a novel approach adopting large language models (LLMs) and feature-generating prompts to address these challenges. We propose a dynamic and adaptive feature generation method that enhances the interpretability of the feature generation process. Our approach broadens the applicability across various data types and tasks and offers advantages in terms of strategic flexibility. A broad range of experiments showcases that our approach is significantly superior to existing methods.
\end{abstract}

\section{Introduction}
The success of machine learning (ML) algorithms generally depends on three aspects: data processing, feature engineering, and modeling~\cite{jordan2015machine}. Among these, feature engineering is critical, directly influencing ML models' performance and effectiveness. Within feature engineering, feature generation is a vital process of transforming raw features into a structured format and optimizing the feature space by creating new features from the original ones~\cite{nam2024optimized}. This transformation usually involves mathematical or algorithmic operations on existing features. This optimization can significantly enrich the feature space, enabling ML models to draw and utilize data information more effectively and achieve superior performance.
\begin{figure}[tbp]
  \centering
  \includegraphics[width=\linewidth]{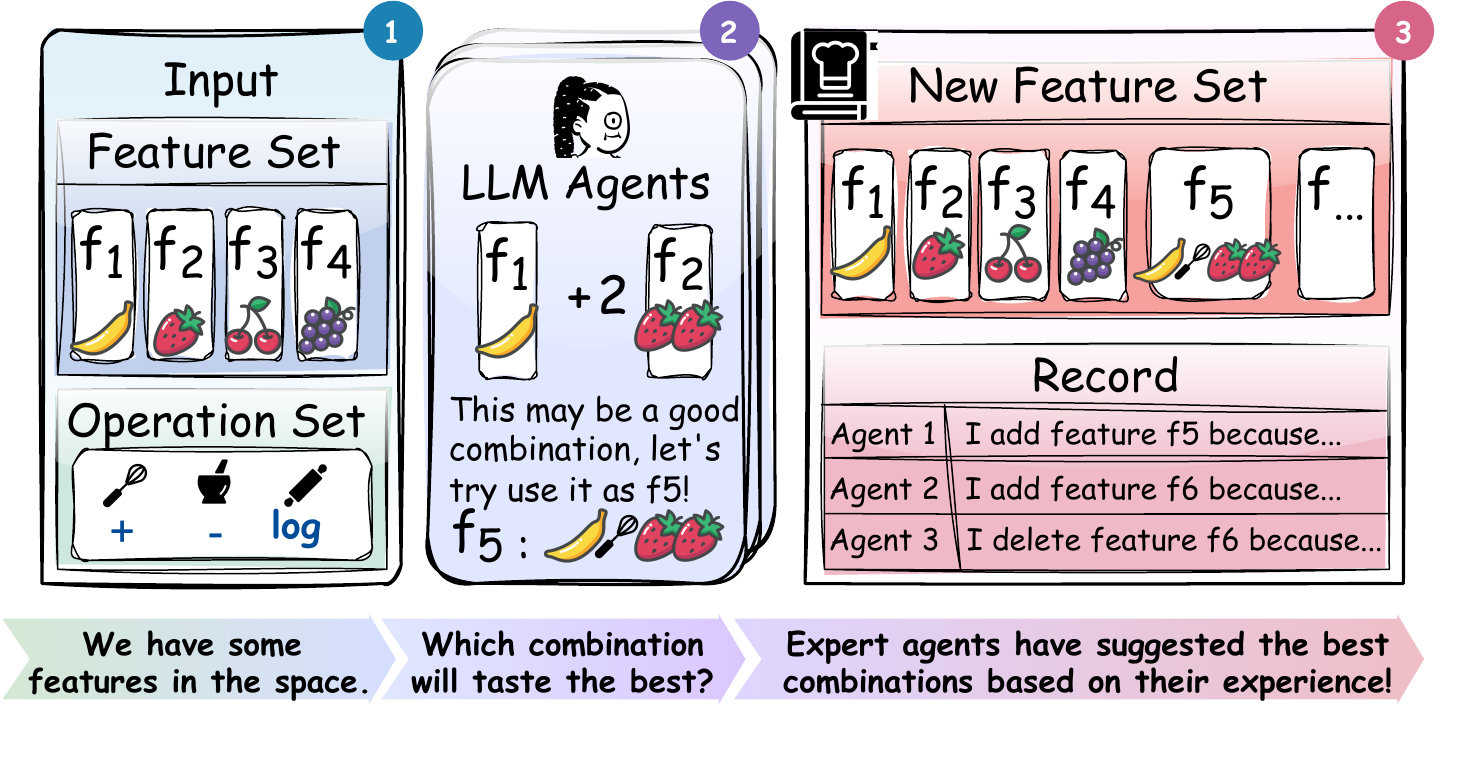}
  \caption{Our goal is to iteratively reconstruct the feature space to create an optimal and explainable feature set that enhances performance on downstream machine learning tasks.}
\end{figure}

Given the importance of feature engineering and feature generation, extensive research has focused on enhancing both the effectiveness and efficiency of these methods~\cite{mumuni2024automated,gu2024large}. As deep learning has explosively developed, automated feature engineering has emerged as a mainstream approach due to its convenience and remarkable performance. These auto-engineering algorithms greatly reduce the need for manual calculation and determination, and thus reduce subjective errors and mistakes. In the field of automated feature
generation, there are outstanding works such as~\cite{khurana2018feature,zhang2024tfwt,wang2022group}. These methods leverage advanced techniques like reinforcement learning to generate optimal feature sets that enhance the performance of downstream tasks while maintaining a degree of explainability. 
Despite these methods enriching the feature space and restructuring it into a more refined space structure, like other automated feature generation and broader automated feature engineering techniques, these works have not resolved three fundamental issues: 1) \textbf{Lack of Explainability:} Most feature engineering processes that utilize deep learning algorithms are result-oriented ``black boxes''. The specific operations on features and the process of restructuring feature space are usually hard for researchers to understand. 2) \textbf{Limited Applicability:} For different data types and downstream tasks, researchers need to manually select and switch among a wide range of feature engineering methods, which demands a high level of domain knowledge and expertise from the researchers. This limited applicability can be a significant barrier across varied ML scenarios. 3) \textbf{Fixed Strategy:} Existing methods usually require predefined strategies for exploration and generation. Models utilizing these methods are unable to adapt or refine their strategies based on insights acquired about the feature space during the model's learning process. Hence, these models lack the flexibility to dynamically adjust to the evolving understanding of the data.

\paragraph{Our Targets.} We aim to develop an automated feature generation method that: (1) offers a \textbf{transparent and understandable} feature generation process; (2) adapts to various data types and downstream tasks by selecting best-suited operations \textbf{automatically}; (3) continuously learns from the feature space and \textbf{refines its strategies} based on new insights. For target (1), we explicitly document the operations applied during the feature generation and how they transform the input data into a structured feature space. By doing so, we intend to replace their lack of interpretability with a more open and interpretable framework. For target (2), we apply a large language model (LLM) that can assess the characteristics of the data and the requirements of the task and dynamically select the most appropriate strategies under proper guidance. For target (3), we develop multiple expert-level LLM agents with different strategies. These agents adjust their operations and strategies in response to evolving data patterns, task requirements, and actions of other agents.

\paragraph{Our Method.} We employ a novel approach to automate and enhance the feature generation process by adopting LLMs with expert-level guidance. Our method starts by inputting an original feature set and a predefined operation set into the LLM, which then instantiates several expert agents. Each agent generates new features with its exclusive strategy and integrates these new features with the original features to produce enriched feature sets. The newly created feature sets are applied to downstream tasks to evaluate their performance, and the performance outcomes are fed back to the agents. The agents ``communicate'' and share their inferences, and reflect on the impact of their own and other agents' operations on the downstream tasks. In this way, the agents refine and optimize their generation strategies to better capture how variations in the feature space impact downstream tasks and, consequently, reconstruct the feature space more adaptively and effectively. The iteration continues with new feature sets until the optimal feature set for the task is found or a predefined number of iterations is reached.

\paragraph{Contributions.}Our main contributions include:
\begin{itemize}

    \item We propose a novel automated feature generation methodology based on LLMs to restructure the feature space of a dataset. This end-to-end structure enables training with LLM agents on varied feature generation strategies without manual selection of feature operations.
    \item We introduce a dynamic and adaptive generation process based on feedback from downstream tasks. This method effectively utilizes the in-context learning and reasoning capabilities of LLMs in the feature generation process. It significantly enhances the applicability and effectiveness of the generated features across various machine learning scenarios without the need to create additional ML models.
    \item We conduct a series of experiments to validate the effectiveness and robustness of our method across different datasets and downstream tasks. Our results demonstrate that our method has clear advantages over existing methods and considerable potential to promote a broader range of feature engineering tasks.
\end{itemize}
\section{Preliminary and Related Work}
\subsection{Feature Generation}
\paragraph{Feature Engineering.} Feature engineering is the process of selecting, modifying, or creating new features from raw data to improve the performance of machine learning models~\cite{severyn2013automatic}. This process can be described as a function $\phi: \mathcal{F} \rightarrow \mathcal{F}^{'}$ that transforms the original feature set $\mathcal{F}$ into a new feature set $\mathcal{F}^{'}$ through certain operations. Feature generation is a decisive category in the field of feature engineering. Feature generation means generating new attributes from existing features in a dataset through various mathematical or logical transformation operations. Feature generation methods can primarily be categorized into two main classes: (1) latent representation learning based methods, such as deep factorization machines ~\cite{guo2017deepfm,song2019autoint} and deep representation learning ~\cite{zhong2016overview,bengio2013representation}. These methods can create complex latent feature spaces; however, their generation processes often lack transparency and are difficult to trace or explain~\cite{scholkopf2021toward}; (2) feature transformation-based methods, which generate new features by arithmetic or aggregation operations~\cite{nargesian2017learning,wang2022semi,wang2022semi2}. These methods often require extensive manual operations and rely heavily on domain knowledge to select transformations and adjust parameters.

\paragraph{Automated Feature Generation.} With the widespread adoption of large models and various deep learning methods in the feature engineering field, automated feature generation is experiencing significant development. Automated feature generation enhances the feature space by systematically creating and integrating new features to improve model performance~\cite{xiang2021physics,wang2022group,pan2020deep,wang2022group}. For example, ~\cite{7837936} developed ExploreKit, which generates a large set of candidate features by combining information from the original features. ~\cite{khurana2016cognito} explores the feature space using manually crafted heuristic traversal strategies, while ~\cite{SHI201881} proposed a feature optimization approach using deep learning and feature selection to enhance traffic classification performance. Although these methods are more efficient than traditional manual feature engineering and enable faster processing of large datasets, they often ignore the semantic aspects of data. Furthermore, their ``black box'' operation process makes the results difficult to explain.

\subsection{LLMs and Tree of Thoughts}
\paragraph{LLMs' Capabilities.} LLMs have revolutionized numerous fields with their extraordinary capabilities in in-context learning and step-by-step reasoning~\cite{xie2024scoring,zhang2024prototypical,wei2022chain}. These capabilities enable LLMs to effectively understand and navigate complex tasks given appropriate guidance and precise prompts. Thus, they exhibit professional competence across various specialized domains~\cite{wang2023aligning,zhang2024retrievalaugmentedfeaturegenerationdomainspecific,zhang2025blindspotnavigationllm,wang2025diversity}. Recent advancements in prompt engineering and reasoning structures have further enabled these models to tackle complex data analysis tasks and multi-step feature engineering processes~\cite{peng2023generating,liu2023summary}.

\paragraph{Prompting.} LLM prompting leverages pre-trained models to recall existing knowledge and generate output based on specific guideline prompts~\cite{white2023prompt}. This process interacts closely with LLMs' capabilities for in-context learning and step-by-step reasoning~\cite{chu2023survey}. Prompts containing context details guide models to think more coherently when handling long tasks. Moreover, step-by-step reasoning prompts can guide a model through logical thought processes by mimicking human-like reasoning to solve complex problems~\cite{yu2023towards,wei2022chain,diao2023active}.

\paragraph{Tree of Thoughts.} One line of work in prompting methods focuses on enhancing the structured reasoning capabilities of LLMs to guide them through logical analysis and problem-solving tasks. \textit{Input-Output (IO) Prompting}~\cite{wang2024survey,hao2023reasoning} involves wrapping the input $x$ with task-specific instructions or examples to guide the model to produce the desired output $y$. This method is straightforward, but its guiding capability is limited in complex tasks. \textit{Chain-of-Thought (CoT) Prompting}~\cite{wei2022chain,besta2024graph,zhang2025leka} involves a sequence of intermediate, coherent language expressions $z_1, \ldots, z_n$ that logically bridge the input $x$ to the output $y$. This method enhances the model's logical capabilities for complex problem-solving, but it is limited to linear, single-strategy reasoning, capturing only a subset of the possible solutions. \textit{Tree of Thoughts (ToT) Prompting}~\cite{yao2024tree} extends the CoT prompting by exploring multiple reasoning paths over thoughts. ToT operates as a search over a tree structure where each node represents a partial solution within the input and contextual thoughts. ToT has advantages over other methods by exploring multiple reasoning paths with different strategies~\cite{zhang2025ratt}. Thus, it performs better in deep data analysis and multi-step feature engineering.
\section{Methodology}
In this section, we design a novel feature generation method called \underline{\textbf{L}}LM \underline{\textbf{F}}eature \underline{\textbf{G}}eneration (\textbf{LFG}), which is highly interpretable, dynamically adaptable, and employs an end-to-end approach to significantly enhance downstream task performance. In this method, we apply an LLM as an auto-generator of features. We augment the LLM with the ToT technique to generate detailed inference along with each step of generation so that each decision is explainable. Furthermore, we collect the performance of downstream tasks to offer feedback to the LLM generator. Our entire workflow is divided into two main parts: (1) \textbf{Automated Feature Generation with LLM Agents}; (2) \textbf{Feedback and Optimization}.

\begin{figure*}[h]
  \centering
  \includegraphics[width=\linewidth]{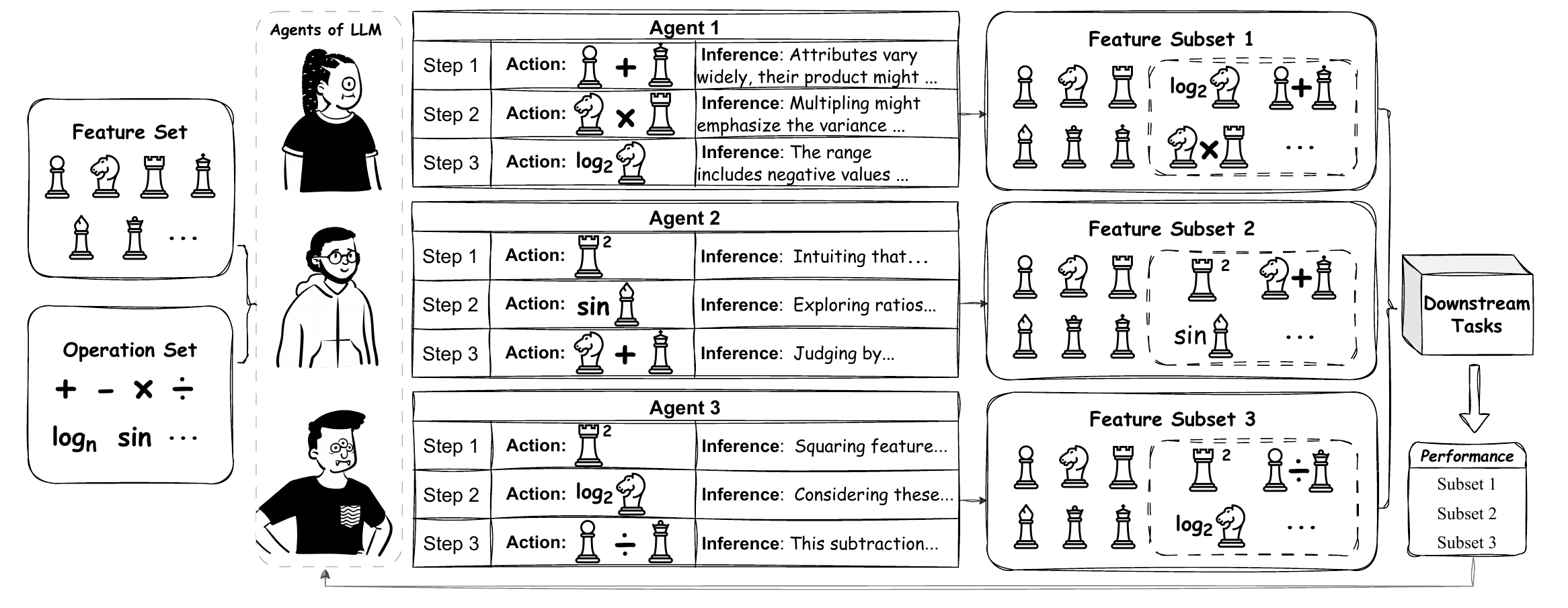}
  \caption{The framework of LFG. First, we input the original feature set and operation set into the LLM's context window. Second, we guide the LLM in creating three expert agents. Each agent generates new features with operations from the operation set, and then combines new features with original features to create a new feature subset. Then, each of these subsets is individually evaluated on a downstream task. After that, we provide the performance of each feature subset in downstream tasks as feedback to the respective agents and iterate the process until the best feature subset is found or the maximum number of iterations is reached.}
\end{figure*}

\subsection{Important Definitions}

We first define the feature set and operation set, along with their corresponding mathematical symbols.

\paragraph{Feature Set.} Let $\mathcal{D} = \{\mathcal{F}, \mathbf{y}\}$ be the dataset. Here, $\mathcal{F} = \{f_1, f_2, \ldots, f_n\}$ is the feature set, where each column represents a feature $f_i$, and each row represents a data sample; $\mathbf{y}$ is the corresponding target label set for the samples. Our main purpose is to reconstruct the feature space of the dataset $\mathcal{D} = \{ \mathcal{F}, \mathbf{y} \}$.

\paragraph{Operation Set.} The operation set $\mathcal{O}$ consists of mathematical operations performed on existing features to generate new ones. Operations are of two types: unary and binary. Unary operations include ``square'', ``exp'', ``log'', etc.; binary operations include ``plus'', ``multiply'', ``divide'', etc.

\paragraph{Optimization Objective.} Our objective is to construct an optimal and interpretable feature space that can enhance the performance of downstream tasks. Given the feature set $\mathcal{F}$ and the operation set $\mathcal{O}$, our optimization goal is to find a reconstructed feature set $\hat{\mathcal{F}}$:

\begin{equation}
    \mathcal{F}^* = \underset{\hat{\mathcal{F}}}{\text{argmax}} \; \mathbf{\theta}_{\mathcal{R}}(\hat{\mathcal{F}}, \mathbf{y}),
\end{equation}

where $\mathcal{R}$ is a downstream ML task (e.g., classification, regression, ranking, detection), $\mathbf{\theta}$ is the performance metric of $\mathcal{R}$, and $\hat{\mathcal{F}}$  is a reconstructed feature set derived from $\mathcal{F}$. Here, $\hat{\mathcal{F}}$ is generated by applying operations from $\mathcal{O}$ to the original feature set $\mathcal{F}$ using a certain algorithmic structure.

\subsection{Automated Feature Generation with LLM Agents}

In this section, we introduce the concept of LLM agents and detail their role in feature generation. We utilize an LLM to generate several agents for feature-generation tasks. Here, each agent acts as an automated feature generator, which applies specific operations on a given feature set to generate new features or delete existing features based on iterative prompts. These agents can emulate expert-level logical reasoning and decision-making capabilities, thus optimizing the feature set for input data of downstream tasks and enhancing their performance.

\paragraph{Agents.} We first define an agent $\mathcal{A}_l$ that operates on a feature subset $\mathcal{F}_l = \{f_1, \ldots, f_n\}$ by applying operations to the features $f_i$ and $f_j$. The $\mathcal{F}_l$ is a subset of $\mathcal{F}$. The agent draws operations from an operation set $\mathcal{O}$, to generate new features:

\begin{equation}
\mathcal{F}_l^{'} = \mathcal{F}_l\cup\{ g_1, \ldots, g_k, \ldots \} .
\end{equation}

Each new feature $g_k$ is produced by applying an operation $o$ to the features $f_i$ and $f_j$:

\begin{equation}
g_k = o(f_i, f_j).
\end{equation}

Here, $\mathcal{F}_l$ is the initial feature subset for $\mathcal{A}_l$; $\mathcal{F}_l^{'}$ is the output feature subset with original and all new features from $\mathcal{A}_l$; and $g_k$ is the new feature from applying operation $o$ to features $f_i, f_j \in \mathcal{F}_l$. Thus, we complete one generation node $s_l$ of $\mathcal{A}_l$, and layer $Layer_1$ consists of $l$ total generation nodes.

\subsection{Feedback and Optimization}

We now have a new feature set composed of all $l$ subsets, denoted as $\mathcal{F}_l^{'} = {\{\mathcal{F}_l, g_1, \ldots, g_k\}}$. 
After the agents generate all $l$ subsets, we move forward to evaluate the performance of these subsets in downstream tasks and the effectiveness of the generated nodes at this layer. We then gather performance metrics from downstream tasks as feedback $\mathbf{\theta} = \{\theta_t\}_{i=1}^T$, for each iteration $t$ corresponding to each generation layer. This feedback reflects the effectiveness of the newly created feature subset and illustrates the agents' contribution to the task. The agents evaluate their performance by comparing the metric $\theta_t$ with $\theta_{(t-1)}$. If $\theta_t$ significantly improves over $\theta_{(t-1)}$, the feature generation strategy is considered ``effective''. The agents' self-evaluation positively correlates with the increase in these metrics; a larger improvement signifies a more successful feature generation strategy.

\par Furthermore, each agent shares its reasoning process during generating features, including the logic behind selecting specific operations. This transparency enables agents to understand and adopt diverse strategies to achieve collaborative improvement. Through self-evaluation and peer interactions, agents identify which strategies successfully enhanced performance and which may require adjustments or replacements. In the subsequent rounds, agents generate increasingly effective feature subsets adopting $\mathcal{F}_l^{'}$ (from the previous round) and $\mathcal{O}$ to form the next layer of generated nodes. The iterative process continues until the optimal feature set $\hat{\mathcal{F}}$ is found or the maximum number of iterations $T$ is reached:

\begin{equation}
\hat{\mathcal{F}} = \underset{\mathcal{F} \in \{\mathcal{F}_l^{\text{opt}} | l \in L\}}{\arg\max} \ \theta(\mathcal{F})
\end{equation}

where $\{\mathcal{F}_l^{\text{opt}} \mid l \in L\}$ is the set of optimal feature subsets from each iteration~$l$, and $\theta(\mathcal{F})$ is the performance score of a feature set~$\mathcal{F}$.


\paragraph{Monte Carlo Tree Search.} In order to further optimize the feature space and balance exploration with exploitation, we employ an improved Monte Carlo Tree Search (MCTS)~\cite{yao2024tree,browne2012survey} to explore new or underutilized feature space structures after the initial $T$ iterations. 
We adopt this MCTS framework because of its random sampling-based search strategy, which enhances the probability of finding superior feature combinations. In our case, this flexible searching method constructs a tree from all agents' generation nodes and adjusts the search direction based on the performance feedback. Here, we define the root node as the original feature set given to the LLM, with each generation node representing a tree node. The search covers all generation nodes in the last iteration layer $Layer_T$ as leaf nodes.

\begin{figure*}[!htbp]
  \centering
  \includegraphics[width=\linewidth]{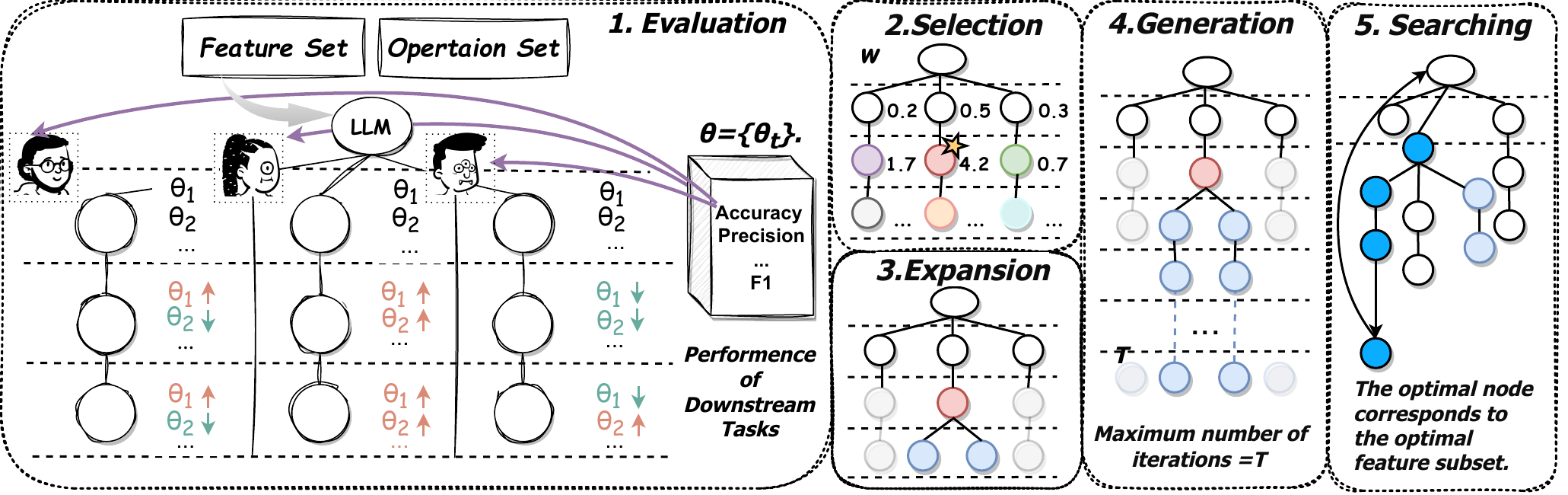}
  \caption{The framework of MCTS for feature generation includes five stages: 1) Performance Evaluation, 2) Node Selection, 3) Node Expansion, 4) Node Generation, 5) Optimal Subset Searching.}
\end{figure*}

We start with node evaluation by calculating the improvement in the downstream task's performance compared to the parent node. The evaluation reflects the effectiveness of newly generated features. Then, the MCTS iteratively builds a search tree by cycling through four phases: selection, expansion, evaluation, and searching. In the selection phase, children nodes are recursively selected from the root utilizing the Upper Confidence Bound (UCB)~\cite{auer2002finite}:
\begin{equation} 
UCB(i) = w_i + C \sqrt{\frac{2 \ln s^{'}_i}{s_i}} ,
\end{equation} 

where $s_i$ and $s^{'}_i$ are the visit counts for node $i$ and its parent respectively, $C$ is a hyperparameter balancing exploration and exploitation. 
The value $w_i$ represents the average performance improvement of all descendant nodes of node $i$, computed based on changes in the downstream task performance metric $\mathbf{\theta}$. Specifically, for each node $i$, the value of $w_i$ is calculated as the average performance improvement observed from its child nodes in the downstream tasks. This is given by:
\begin{equation}
w_i = \frac{1}{|\mathcal{C}_i|} \sum_{j \in \mathcal{C}_i} (\theta_t^{(j)} - \theta_{(t-1)}^{(j)}) .
\end{equation}
Here, $\mathcal{C}_i$ is the set of child nodes of node $i$, and $|\mathcal{C}_i|$ is its cardinality (i.e., the number of child nodes of $i$). $\Delta \theta^{(j)} = (\theta_t^{(j)} - \theta_{(t-1)}^{(j)})$ represents the performance improvement contributed by child node $j \in \mathcal{C}_i$, where $\theta_t^{(j)}$ and $\theta_{(t-1)}^{(j)}$ are the performance metrics associated with the state after exploring child $j$ and the state before, respectively.

In the tree, we select the most promising node $s_l$. We define the selection criterion for each node as:     
\begin{equation}     
s_l^{\text{select}} = \argmax_{s_l \in L} \left\{ \phi(s_l) + \psi(s_l) \right\},     
\end{equation}   

where $\psi(s_l)$ and $\phi(s_l)$ represent the current and future value of the node. We apply further strategies to generate and expand feature subsets upon selecting these nodes.

On the selected nodes $s_l^{\text{select}}$, more complex or novel feature operations are implemented to generate additional feature subsets. These newly generated subsets, $\mathcal{F}_l^{'}$, are then used for further evaluation in downstream tasks to validate their efficacy. The process is mathematically defined as:     
\begin{equation}     
\mathcal{F}_l' = \mathcal{G}(\mathcal{F}_l^{\text{select}}, \mathcal{O}'),   
\end{equation}    

where $\mathcal{G}$ is the transformation function generating $\mathcal{F}_l^{'}$ by applying operations from a selected subset $\mathcal{O}' \subseteq \mathcal{O}$ (the predefined complete operation set) to $\mathcal{F}_l^{\text{select}}$. At the end of the MCTS process, the optimal feature subset is determined by choosing the leaf node with the highest $w_i$ value, representing the feature set that maximized performance improvement during simulations:
\begin{equation}
\mathcal{F}^* = \mathcal{F}_{i^*}, \quad \text{where} \quad i^* = \underset{i \in \text{LeafNodes}}{\arg\max} \; w_i.
\end{equation}
This $\mathcal{F}^*$ represents the optimal feature subset identified by the MCTS, which should then be validated further in real-world tasks to confirm its efficacy.

\section{Experiments}

In this section, we present three experiments to demonstrate the effectiveness of the \model. First, we compare \model against several baseline methods on multiple downstream classification tasks. Second, we perform a robustness check on \model's performance improvement. Finally, we further study and analyze iterative performance improvements in experimental results and discuss their reasons.

\subsection{Experimental Setup}
\paragraph{Datasets.} We evaluate the \model method on four real-world datasets from UCI, including \textit{Ionosphere (Ino)}~\cite{misc_ionosphere_52}, \textit{Amazon Commerce Reviews (Ama)}~\cite{misc_amazon_commerce_reviews_215}, and \textit{Abalone (Aba)}
~\cite{misc_abalone_1}, as well as \textit{Diabetes Health Indicators Dataset (Dia)}~\cite{teboul2022diabetes} from Kaggle. The detailed information is shown in Table~\ref{tab:dataset}. For each dataset, we randomly selected \(55\%\) of the data for training.

\begin{table}[tbp]
\centering
\fontsize{9}{11}\selectfont
\setlength{\tabcolsep}{8pt}

\begin{tabular}{l|lll}
    \toprule
    Datasets & Samples & Features & Class \\
    \midrule
    Ion    & 351      & 34     & 2 \\
    Ama    & 1,500    & 10,000 & 2 \\
    Aba    & 4,177    & 8      & 3 \\
    Dia    & 441,455  & 330    & 2 \\
    \bottomrule
\end{tabular}
\caption{Datasets descriptions.}
\label{tab:dataset}
\end{table}

\paragraph{Metrics.} We evaluate the model performance by the following metrics: \textit{Overall Accuracy (Acc)} measures the proportion of true results (both true positives and true negatives) in the total dataset. \textit{Precision (Prec)} reflects the ratio of true positive predictions to all positive predictions for each class. \textit{Recall (Rec)}, also known as sensitivity, reflects the ratio of true positive predictions to all actual positives for each class. \textit{F-Measure (F1)} is the harmonic mean of precision and recall, providing a single score that balances both metrics. 

\paragraph{Downstream Tasks.} We apply the \model model across a range of classification models, including \textit{Random Forests (RF)}, \textit{Decision Tree (DT)}, \textit{K-Nearest Neighbor (KNN)} and \textit{Multilayer Perceptrons (MLP)}. We compare the performance outcomes in these models both with and without our method.

\paragraph{Baseline Models.} We compare the \model method with raw data (Raw), the Least Absolute Shrinkage and Selection Operator (Lasso), and Feature Engineering for Predictive Modeling using Reinforcement Learning~\cite{khurana2018feature} (RL). Here, we set the same operation set for both \model and RL method to consist of \textit{square root, square, cosine, sine, tangent, exp, cube, log, reciprocal, sigmoid, plus, subtract, multiply}, and \textit{divide}. For our LLM, we perform all the experiments on the OpenAI API, GPT-3.5 Turbo model~\cite{openai_api}.

\subsection{Experimental Results}
\paragraph{Overall Performance.} Table~\ref{tab:addlabel} shows the overall performance results for \model and the baseline models.

(1) Comparison with baseline models. From the table, we can see that the \model method consistently surpasses baseline methods across a variety of metrics and datasets. Specifically, we show the results of \model in 3 iterations of generation, denoted as \model-3,  and compare it with the full \model in \(T \leq 10\). On dataset \textit{Ion}, the \model achieves the best accuracy of \(95.6\%\), which is a \(6.4\%\) increase over raw data on task RF, and \(4.5\%\) over the RL method. On dataset \textit{Ama}, the \model-3 improves the accuracy of raw data of \(59.6\%\) to \(61.5\%\) while \model performs an increase of \(4.3\%\) on RF. Beyond accuracy, \model consistently demonstrates significant improvements over baseline methods across other metrics and datasets. For instance, on dataset \textit{Ion}, the highest recall improvement of 9.8\% on DT reflects \model's ability to effectively handle positive samples. Similarly, the improvements in F1, such as increasing from 0.476 to 0.607 on dataset \textit{Ama} on KNN and from 0.855 to 0.932 on dataset \textit{Ion} on DT, showcase the model's adaptability across diverse tasks and feature distributions.

(2) Performance across different models and metrics. In precision, the \model also shows superior performance. For example, on dataset \textit{Ion}, the \model reaches a precision of \(90.6\%\), surpassing the highest precision of \(86.7\%\) among all baselines on KNN. This result indicates that the \model is effective in reducing misclassifications. Regarding the metric recall, the \model increases the \textit{Ion}'s recall by \(9.8\%\) compared to vanilla data and by \(4.3\%\) compared to best of baselines on DT task, which demonstrates that \model
brings higher improvement on identifying positive samples. On the \textit{Ama} dataset, our method significantly improves F1. The 3-iteration version, \model-3, increased the F1 from an initial \(63.8\%\) to \(67.6\%\). The full \model model further boosted this performance, reaching a final F1 of \(68.0\%\). As the harmonic mean of precision and recall, this substantial improvement in F1 showcases the model's capability to effectively balance between reducing misclassifications and minimizing missed classifications.

\begin{table*}[tbp]
  \centering

  \resizebox{0.98\textwidth}{!}{
    \begin{tabular}{|c|l|cccc|cccc|cccc|cccc|}
    \hline
    \multirow{3}[2]{*}{\textbf{Metrics}} & \multicolumn{1}{c|}{\multirow{3}[2]{*}{\textbf{Model}}} & \multicolumn{4}{c|}{\multirow{3}[2]{*}{\textbf{RF}}} & \multicolumn{4}{c|}{\multirow{3}[2]{*}{\textbf{DT}}} & \multicolumn{4}{c|}{\multirow{3}[2]{*}{\textbf{KNN}}} & \multicolumn{4}{c|}{\multirow{3}[2]{*}{\textbf{MLP}}} \\
          &       & \multicolumn{4}{c|}{}         & \multicolumn{4}{c|}{}         & \multicolumn{4}{c|}{}         & \multicolumn{4}{c|}{} \\
          &       & \multicolumn{4}{c|}{}         & \multicolumn{4}{c|}{}         & \multicolumn{4}{c|}{}         & \multicolumn{4}{c|}{} \\
    \hline
    \multirow{6}[11]{*}[3ex]{\textbf{Acc}} & \rule{0pt}{10pt} & \multicolumn{1}{c}{Ion} & \multicolumn{1}{c}{Ama} & \multicolumn{1}{c}{Aba} & \multicolumn{1}{c|}{Dia} & \multicolumn{1}{c}{Ion} & \multicolumn{1}{c}{Ama} & \multicolumn{1}{c}{Aba} & \multicolumn{1}{c|}{Dia} & \multicolumn{1}{c}{Ion} & \multicolumn{1}{c}{Ama} & \multicolumn{1}{c}{Aba} & \multicolumn{1}{c|}{Dia} & \multicolumn{1}{c}{Ion} & \multicolumn{1}{c}{Ama} & \multicolumn{1}{c}{Aba} & \multicolumn{1}{c|}{Dia} \\
\cline{2-18}          & Raw \rule{0pt}{10pt}  & 0.892  & 0.596  & 0.539  & 0.859  & 0.873  & 0.536  & 0.489  & 0.794  & 0.810  & 0.548  & 0.529  & 0.847  & 0.892  & 0.639  & 0.546  & 0.862  \\
\cline{2-2}          & Lasso \rule{0pt}{10pt} & 0.892  & \underline{0.627}  & 0.537  & \underline{0.860}  & 0.880  & 0.554  & 0.491  & 0.796  & 0.848  & 0.533  & 0.532  & 0.849  & 0.905  & 0.671  & 0.552  & 0.863  \\
\cline{2-2}          & RL \rule{0pt}{10pt}    & 0.911  & 0.607  & 0.559  & 0.859  & 0.892  & 0.548  & 0.508  & 0.796  & 0.829  & 0.542  & 0.534  & 0.848  & 0.911  & 0.616  & 0.558  & 0.864  \\
\cline{2-2}          & LFG-3 \rule{0pt}{10pt} & \underline{0.918}  & 0.615  & \underline{0.564}  & \underline{0.860}  & \underline{0.905}  & \underline{0.590}  & \underline{0.523}  & \underline{0.800}  & \underline{0.873}  & \underline{0.607} & \underline{0.541}  & \underline{0.848}  & \underline{0.918}  & \underline{0.676}  & \underline{0.572}  & \underline{0.866}  \\
\cline{2-2}          & LFG \rule{0pt}{10pt}  & \textbf{0.956 } & \textbf{0.639 } & \textbf{0.564}  & \textbf{0.861}  & \textbf{0.937 } & \textbf{0.590}  & \textbf{0.523}  & \textbf{0.800}  & \textbf{0.886 } & \textbf{0.607 } & \textbf{0.541}  & \textbf{0.848}  & \textbf{0.943 } & \textbf{0.680 } &  \textbf{0.573}  & \textbf{0.866}  \\
\hline
    \multirow{6}[10]{*}[3ex]{\textbf{Prec}} & \rule{0pt}{10pt} & \multicolumn{1}{c}{Ion} & \multicolumn{1}{c}{Ama} & \multicolumn{1}{c}{Aba} & \multicolumn{1}{c|}{Dia} & \multicolumn{1}{c}{Ion} & \multicolumn{1}{c}{Ama} & \multicolumn{1}{c}{Aba} & \multicolumn{1}{c|}{Dia} & \multicolumn{1}{c}{Ion} & \multicolumn{1}{c}{Ama} & \multicolumn{1}{c}{Aba} & \multicolumn{1}{c|}{Dia} & \multicolumn{1}{c}{Ion} & \multicolumn{1}{c}{Ama} & \multicolumn{1}{c}{Aba} & \multicolumn{1}{c|}{Dia} \\
\cline{2-18}          & Raw \rule{0pt}{10pt}  & 0.884  & 0.598  & 0.533  & 0.681  & \underline{0.899}  & 0.537  & 0.492  & 0.588  & 0.867  & 0.605  & 0.525  & 0.643  & 0.900  & 0.638  & 0.541  & 0.697  \\
\cline{2-2}          & Lasso \rule{0pt}{10pt} & 0.866  & 0.624  & 0.531  & 0.684  & 0.857  & 0.554  & 0.493  & 0.590  & 0.865  & 0.544  & 0.532  & 0.647  & 0.904  & 0.671  & 0.549  & 0.703  \\
\cline{2-2}          & RL  \rule{0pt}{10pt}  & \underline{0.919}  & 0.610  & 0.554  & 0.683  & 0.892  & 0.548  & 0.495  & 0.590  & 0.856  & 0.542 & 0.528  & 0.642  & 0.915  & 0.616  & 0.544  & 0.708  \\
\cline{2-2}          & LFG-3 \rule{0pt}{10pt} & 0.915  & \underline{0.623}  & \underline{0.562}  & \underline{0.688}  & 0.892  & \underline{0.591}  & \underline{0.524}  & \underline{0.592}  & \underline{0.891}  & \underline{0.608}  & \underline{0.544}  & \underline{0.645}  & \underline{0.940}  & \underline{0.676}  & \underline{0.560}  & \underline{0.719}  \\
\cline{2-2}          & LFG \rule{0pt}{10pt}  & \textbf{0.962}  & \textbf{0.639}  & \textbf{0.562}  & \textbf{0.690}  & \textbf{0.929}  & \textbf{0.591}  & \textbf{0.524}  & \textbf{0.592}  & \textbf{0.906}  & \textbf{0.608}  & \textbf{0.544}  & \textbf{0.645}  & \textbf{0.953}  & \textbf{0.681}  & \textbf{0.562}  & \textbf{0.719}  \\
\hline
    \multirow{6}[11]{*}[3ex]{\textbf{Rec}} & \rule{0pt}{10pt} & \multicolumn{1}{c}{Ion} & \multicolumn{1}{c}{Ama} & \multicolumn{1}{c}{Aba} & \multicolumn{1}{c|}{Dia} & \multicolumn{1}{c}{Ion} & \multicolumn{1}{c}{Ama} & \multicolumn{1}{c}{Aba} & \multicolumn{1}{c|}{Dia} & \multicolumn{1}{c}{Ion} & \multicolumn{1}{c}{Ama} & \multicolumn{1}{c}{Aba} & \multicolumn{1}{c|}{Dia} & \multicolumn{1}{c}{Ion} & \multicolumn{1}{c}{Ama} & \multicolumn{1}{c}{Aba} & \multicolumn{1}{c|}{Dia} \\
\cline{2-18}          & Raw \rule{0pt}{10pt}  & 0.880  & 0.597  & 0.545  & 0.570  & 0.837  & 0.537  & 0.493  & 0.597  & 0.768  & 0.548  & 0.533  & 0.577  & 0.892  & 0.639  & 0.544  & 0.595  \\
\cline{2-2}          & Lasso \rule{0pt}{10pt} & 0.887  & \underline{0.622}  & 0.541  & \underline{0.575}  & 0.874  & 0.554  & 0.496  & 0.599  & 0.802  & 0.510  & 0.542  & 0.580  & 0.885  & 0.672  & 0.552  & 0.591  \\
\cline{2-2}          & RL \rule{0pt}{10pt}   & 0.911  & 0.607  & 0.559  & 0.568  & 0.892  & 0.548  & 0.508  & 0.598  & \underline{0.829}  & 0.542  & 0.534  & 0.579  & \underline{0.911}  & 0.616  & 0.558  & \textbf{0.589}  \\
\cline{2-2}          & LFG-3 \rule{0pt}{10pt} & \underline{0.904}  & 0.618  & \underline{0.564}  & \underline{0.568}  & \underline{0.902}  & \underline{0.591}  & \underline{0.526}  & \underline{0.601}  & 0.828  & \underline{0.607}  & \underline{0.546}  & \underline{0.582}  & 0.897  & \underline{0.676}  & \underline{0.573}  & \underline{0.562}  \\
\cline{2-2}         & LFG \rule{0pt}{10pt}  & \textbf{0.944}  & \textbf{0.639}  & \textbf{0.564}  & \textbf{0.570}  & \textbf{0.935}  & \textbf{0.591}  & \textbf{0.526}  & \textbf{0.601}  & \textbf{0.828}  & \textbf{0.607}  & \textbf{0.546}  & \textbf{0.582}  & \textbf{0.921}  & \textbf{0.681}  & \textbf{0.565}  & \underline{0.562}  \\
\hline    \multirow{6}[12]{*}[3ex]{\textbf{F1}} & \rule{0pt}{10pt} & \multicolumn{1}{c}{Ion} & \multicolumn{1}{c}{Ama} & \multicolumn{1}{c}{Aba} & \multicolumn{1}{c|}{Dia} & \multicolumn{1}{c}{Ion} & \multicolumn{1}{c}{Ama} & \multicolumn{1}{c}{Aba} & \multicolumn{1}{c|}{Dia} & \multicolumn{1}{c}{Ion} & \multicolumn{1}{c}{Ama} & \multicolumn{1}{c}{Aba} & \multicolumn{1}{c|}{Dia} & \multicolumn{1}{c}{Ion} & \multicolumn{1}{c}{Ama} & \multicolumn{1}{c}{Aba} & \multicolumn{1}{c|}{Dia} \\
\cline{2-18}          & Raw \rule{0pt}{10pt}  & 0.882  & 0.595  & 0.537  & \underline{0.587}  & 0.855  & 0.536  & 0.492  & 0.592  & 0.781  & 0.476  & 0.528  & 0.592  & 0.889  & 0.638  & 0.541  & \textbf{0.618}  \\
\cline{2-2}          & Lasso \rule{0pt}{10pt} & 0.875  & \underline{0.622}  & 0.533  & \textbf{0.594}  & 0.865  & 0.554 & 0.493  & 0.594  & 0.821  & 0.403  & 0.534  & 0.595  & 0.893  & 0.671  & 0.550  & \underline{0.614}  \\
\cline{2-2}          & RL \rule{0pt}{10pt}   & 0.909  & 0.602  & 0.556  & 0.584  & 0.890  & 0.548  & 0.498  & 0.594  & 0.814  & 0.542  & 0.529  & 0.594  & 0.909  & 0.615  & 0.544  & 0.611  \\
\cline{2-2}          & LFG-3 \rule{0pt}{10pt} & \underline{0.909}  & 0.612  & \underline{0.562}  & 0.584  & \underline{0.897}  & \underline{0.590}  & \underline{0.525}  & \underline{0.596}  & \underline{0.849}  & \underline{0.607}  & \underline{0.541}  & \underline{0.598}  & \underline{0.911}  & \underline{0.676}  & \underline{0.563}  & 0.576  \\
\cline{2-2}          & LFG \rule{0pt}{10pt}  & \textbf{0.952}  & \textbf{0.639}  & \textbf{0.562}  & \underline{0.587}  & \textbf{0.932}  & \textbf{0.590}  & \textbf{0.525}  & \textbf{0.596}  & \textbf{0.854} & \textbf{0.607}  & \textbf{0.541}  & \textbf{0.598}  & \textbf{0.935}  & \textbf{0.680}  & \textbf{0.563}  & 0.576  \\
    \bottomrule
    \end{tabular}%
    }
  
  \caption{Overall performance on downstream tasks. The best results are highlighted in \textbf{bold}, and the runner-up results are highlighted in \underline{underline}. (Higher values indicate better performance.)}
  \label{tab:addlabel}
  
\end{table*}

\paragraph{Robustness Check.} We validate the experiments through five-fold cross-validation. The \model method demonstrates robust performance, achieving notable metric improvements. With \model-3, it gains \(2.72\%\) in accuracy and approximately \(5\%\) in precision, recall, and F1. The full \model\ model further boosts these gains to \(4.81\%\) in accuracy and approximately \(7\%\) in other metrics. This consistent enhancement is driven by the extended iterations, which continue to refine the feature space, and it illustrates the adaptability, effectiveness, and reliability of the \model\ method.

\begin{figure}[tbp]
  \centering
  
  \includegraphics[width=1\linewidth]{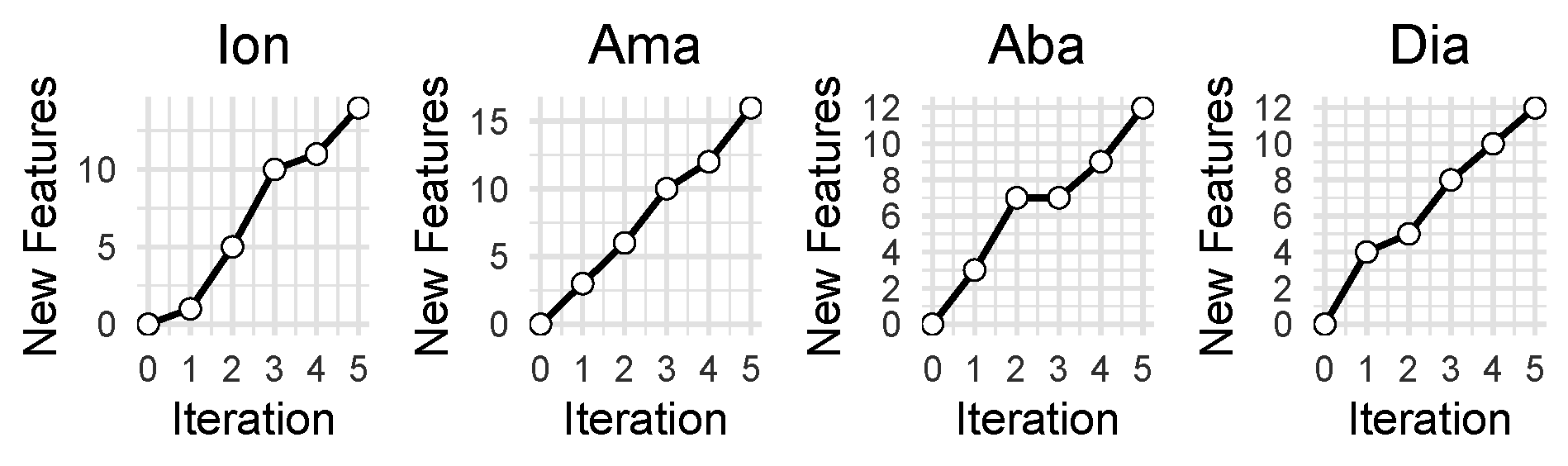}
  \caption{Increase of feature numbers.}
  \label{fig:feautre}
   
\end{figure}

\paragraph{Iterative Performance Improvements.} We then compare the performance improvements of \model-3 and \model. On one hand, the incremental improvement from \model-3 to \model illustrates the continuous enhancement in all metrics, indicating that during additional iterations, the model refines and adjusts its generation strategy to explore better solutions for the feature space. On the other hand, the results of \model-3 with only three iterations already show notable enhancement in all datasets. This shows that the model effectively and rapidly finds the zones of optimal feature space within a limited number of steps. As shown in Figure~\ref{fig:feautre}, we can see that the average number of features in a feature subset increases continuously in the first \(5\) rounds of generation. The increase in features implies that the feature generation algorithm is actively finding new,  potential feature combinations in the feature space, effectively expanding the feature space.

\begin{figure}[!t]
  \centering
  \includegraphics[width=0.95\linewidth]{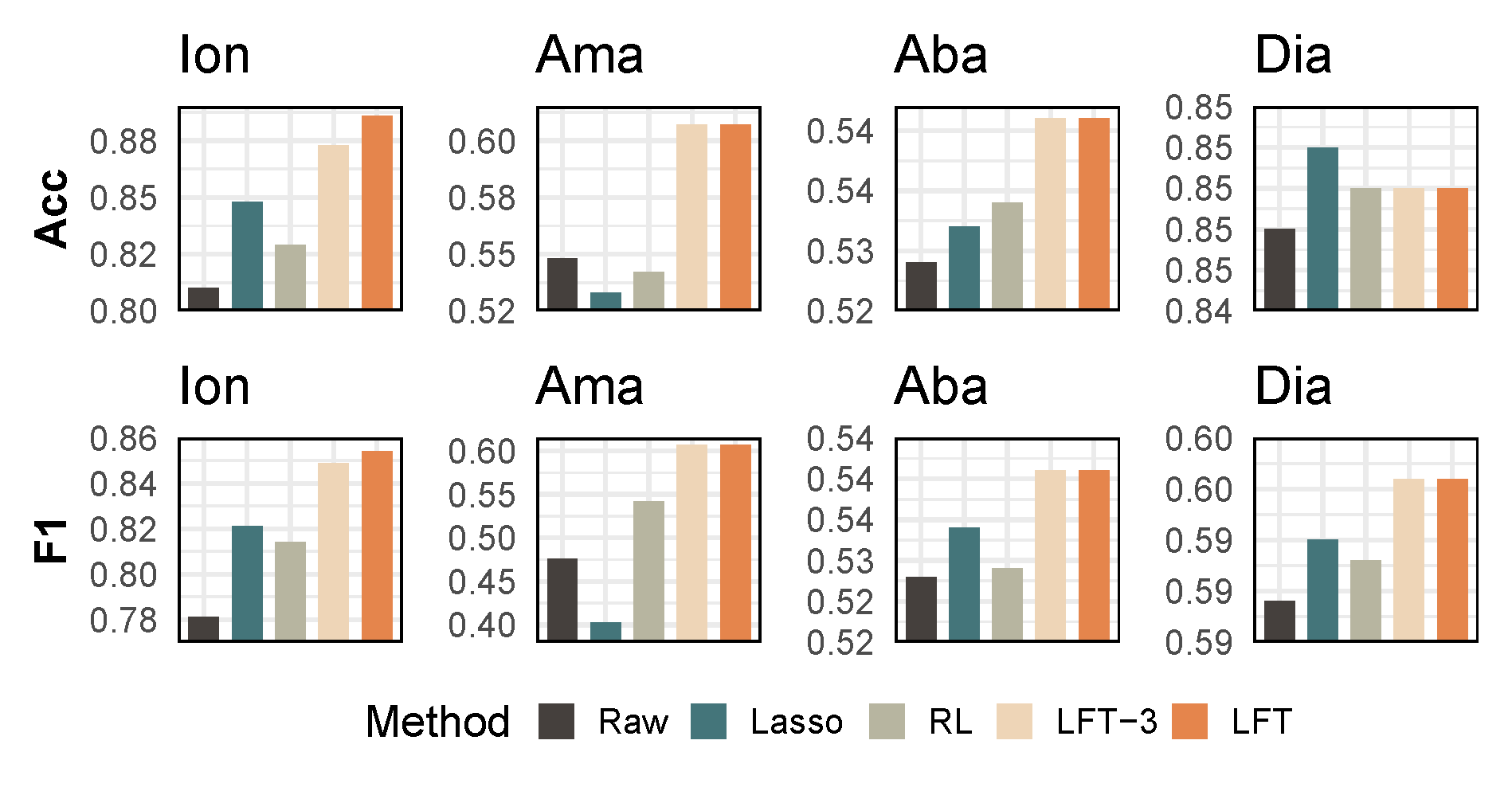}
  \caption{Comparison on KNN (Accuracy and F1).}
  \label{fig:KNN-mini}
  
\end{figure}

\begin{figure}[htbp]
  \centering
  
  \includegraphics[width=0.98\linewidth]{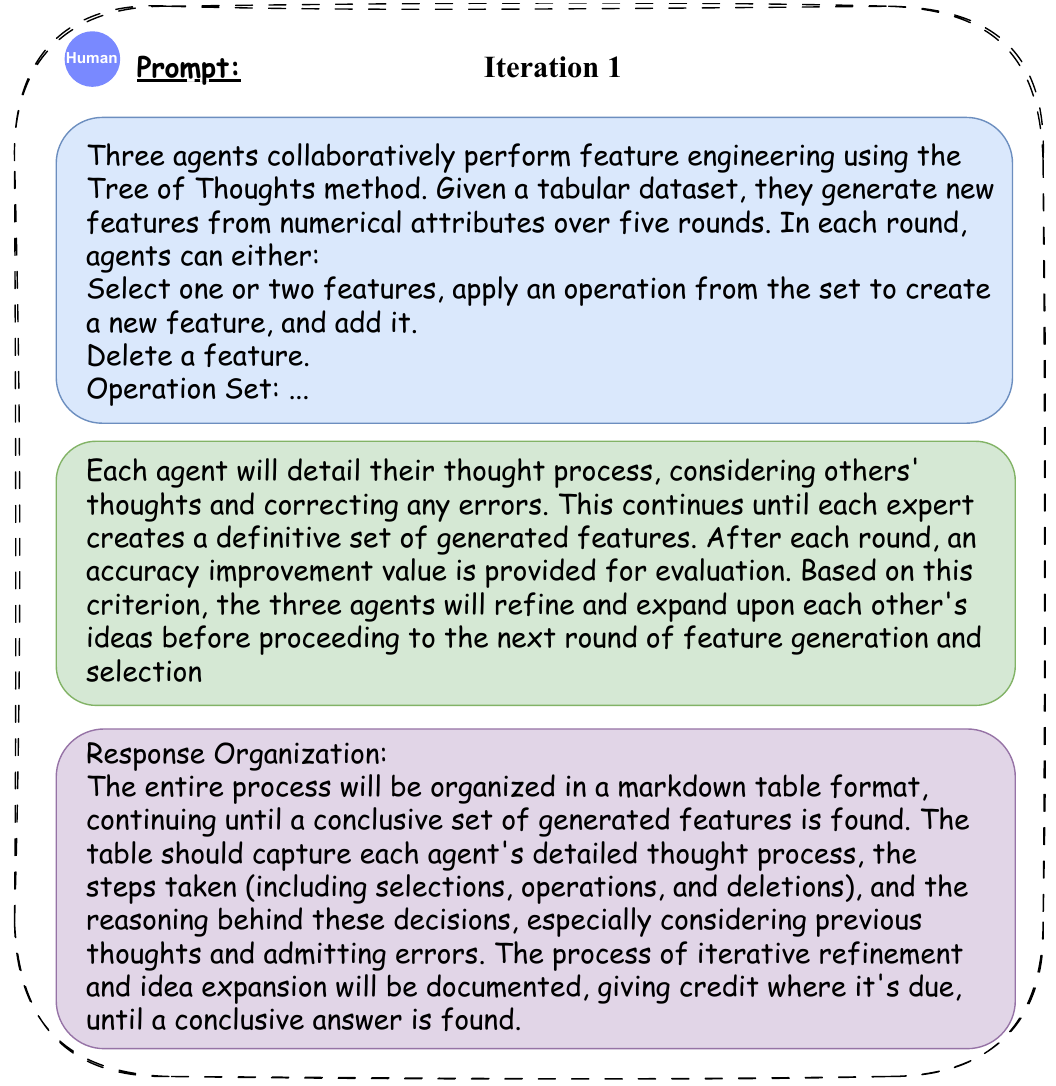}
  \caption{The exemplary prompt for LLM to generate agents and features. The \textit{Feature Engineering} is shown in \colorbox{bgcolor1}{blue}, the \textit{Iterative Refinement and Evaluation} is shown in \colorbox{bgcolor2}{green} and the \textit{Markdown and Organization} is shown in \colorbox{bgcolor3}{purple}. The LLM follows the instructions to start the first iteration and continues to generate features with the downstream tasks' performance data.}
\end{figure}

\section{Conclusion}
In this paper, we present a novel feature generation method \model utilizing LLMs to enhance automated feature engineering's interpretability, adaptability, and strategic flexibility. Our target in designing the model is to address traditional feature generation challenges, including a lack of explainability, limited applicability, and rigid strategy formulation. Our approach utilizes expert-level LLM agents to overcome these. These problems can significantly limit the broader deployment of feature engineering on machine learning tasks in diverse scenarios. Thus, we present \model with a dynamic, adaptive, automated feature generation approach to enhance interpretability and extend utility across various data types and tasks. Moreover, we present extensive experimental results demonstrating that our approach outperforms various existing methods. For future work, we plan to extend this automated feature engineering paradigm across various techniques and machine learning tasks and investigate transformations our approach applies to the feature space to gain deeper insights into its underlying mechanisms and impacts.

\section{Limitations and Ethics Statements}
While our \model shows significant advancements and wide adaptability, there are several limitations that require further exploration, including: (1) high computational demands and limited scalability with very large or complex datasets. With LLM agents being the generator, the computational demands are comparatively higher than traditional methods; (2) the effectiveness of the generated features heavily relies on the quality of input data, which can affect the model's performance in scenarios with poorly curated or noisy datasets; (3) the current work focuses solely on the tabular feature generation process. More complex generation methods on different data are not considered. Extending the \model to handle feature generation in non-tabular scenarios remains a challenge.

The \model framework can enhance the effectiveness and explainability of the feature generation process. However, as the framework uses pre-trained GPT-3.5 Turbo as the generation model, it may inherit the ethical concerns associated with GPT-3.5 Turbo, such as responding to harmful queries or exhibiting biased behaviors.

\newpage
\clearpage

\section*{Acknowledgements}

Pengfei Wang is supported by the National Natural Science Foundation of China (Grant Nos. 62406306 and 92470204). 

\bibliographystyle{named}
\bibliography{ijcai25}

\end{document}